\documentclass[10pt]{cai}

\usepackage{times}
\usepackage{latexsym}

\usepackage[T1]{fontenc}

\usepackage[utf8]{inputenc}

\usepackage{microtype}

\usepackage{inconsolata}

\usepackage{graphicx}

\usepackage{algorithm}
\usepackage{algpseudocode}
\usepackage{graphicx} 
\usepackage{multirow} 
\usepackage{adjustbox}
\usepackage[dvipsnames]{xcolor}
\usepackage{subcaption}
\usepackage{multicol}
\usepackage{pgfplots}
\usepackage{tikz,ifthen}
\usepackage{amsmath} 
\usepackage{appendix}

\usetikzlibrary{shapes.geometric, arrows}

\tikzstyle{process} = [rectangle, minimum width=3cm, minimum height=1cm, text centered, draw=black]
\tikzstyle{decision} = [diamond, minimum width=3cm, minimum height=1cm, text centered, draw=black]
\tikzstyle{arrow} = [thick,->,>=stealth]
\usetikzlibrary{shapes.geometric, arrows}
\tikzstyle{process} = [rectangle, minimum width=3cm, minimum height=1cm, text centered, draw=black]
\tikzstyle{decision} = [diamond, minimum width=3cm, minimum height=1cm, text centered, draw=black]
\tikzstyle{arrow} = [thick,->,>=stealth]
\usetikzlibrary{shapes.geometric, arrows, calc, arrows.meta}
\usetikzlibrary{calc} 

\definecolor{bblue}{HTML}{4F81BD}
\definecolor{rred}{HTML}{C0504D}
\definecolor{ggreen}{HTML}{9BBB59}
\definecolor{ppurple}{HTML}{9F4C7C}

\begin{document}
\def\conferenceyear{2026}
\begin{center}

\title{EPAS: Efficient Training with Progressive Activation Sharing}
\footnote{This is a preprint of a paper accepted at the 39th Canadian Conference on Artificial Intelligence (Canadian AI 2026).}
\maketitle

\thispagestyle{empty}


\begin{tabular}{cc}
Rezaul Karim\upstairs{\affilone,*}, Maryam Dialameh\upstairs{\affilone \affiltwo}, Yang Liu\upstairs{\affilone}, Boxing Chen\upstairs{\affilone}, Walid Ahmed\upstairs{\affilone}
\\[0.25ex]
{\small \upstairs{\affilone} Ascend Team, Huawei Technologies, Toronto, Canada} \\
{\small \upstairs{\affiltwo} Department of Mechanical and Mechatronics Engineering, University of Waterloo, Canada} \\
\end{tabular}
  
\emails{
  \upstairs{*}rezaul.karim3@huawei.com 
}
\vspace*{0.2in}
\end{center}

\begin{abstract}
We present a novel method for \textbf{E}fficient training with \textbf{P}rogressive \textbf{A}ctivation \textbf{S}haring (EPAS). This method bridges progressive training paradigm with the phenomenon of redundant $QK$ (or $KV$) activations across deeper layers of transformers. EPAS gradually grows a sharing region during training by switching decoder layers to activation sharing mode. This results in throughput increase due to reduced compute. To utilize deeper layer redundancy, the sharing region starts from the deep end of the model and grows towards the shallow end. The EPAS trained models allow for variable region lengths of activation sharing for different compute budgets during inference. Empirical evaluations with $QK$ activation sharing in LLaMA models ranging from 125M to 7B parameters show  up to an 11.1\% improvement in training throughput and up to a 29\% improvement in inference throughput while maintaining similar loss curve to the baseline models. Furthermore, applying EPAS in continual pretraining to transform TinyLLaMA into an attention-sharing model yields up to a 10\% improvement in average accuracy over state-of-the-art methods, emphasizing the significance of progressive training in cross layer activation sharing models.

\end{abstract}


\begin{keywords}{Keywords:}
Low Resource NLP, LLM efficiency, efficient inference, parameter-efficient-training, Many-in-one model.
\end{keywords}

\section{Introduction}

Recent research in computational efficiency of Transformers has focused on efficient pretraining~\cite{he2024softdedup}, continual learning~\cite{zhao2024sapt}, fine-tuning~\cite{zhang2024quantized,liu2024aflora,ge2024time} and inference~\cite{alizadeh2024llm,Wu2024LayerCondensedKC}. However, holistic approaches to efficient training and inference remains underexplored as evident from large accuracy-efficiency trade-offs in this direction~\cite{kim2024sparseflow,song2024sleb}. Surprisingly, large transformer models are found to compute redundant activations across deeper layers~\cite{liao2024beyond,brandon2024reducing,hooper2024kvquant,tomar2025xquant,qiao2025swiftkv}. Therefore, as a promising yet less explored direction, we focus on utilizing redundancy  phenomenon towards an unified efficiency solution to training and inference with minimal tradeoffs to accuracy. 

Deeper layers of transformer models have been found to exhibit redundancy in activations of the attention block across layers. For example, multiple deeper layers are found to compute mostly similar attention scores \cite{liao2024beyond,ying2021lazyformer,mu2024cross}. Hence, recent methods reuse $QK$ or $KV$ activations across layers to enhance computational efficiency.
These approaches are generally known as activation sharing.
Attention sharing approaches compute only the value ($V$) and reuse the computed attention score directly from a previous layer to make the model compute efficient~\cite{liao2024beyond,ying2021lazyformer,mu2024cross}. Since the attention score from the previous layer is not directly available in block factoring-based efficient attention algorithms, such as Flash-Attention~\cite{dao2022flashattention,dao2023flashattention}, an alternative approach shares $QK$ across layers~\cite{rajabzadeh2024echoatt}. Meanwhile, some other approaches have proposed to compute the query ($Q$) and reuse ($KV$) from a previous layer~\cite{brandon2024reducing,sun2024you}. The sharing of $QK$ offers greater computational savings, while sharing $KV$ has greater impact in reducing inference memory footprint. 
 
State-of-the-art activation sharing approaches enhance efficiency mostly by sharing activations across the layers of trained models during the inference~\cite{liao2024beyond} or follow model distillation~\cite{rajabzadeh2024echoatt}. Few models incorporate efficient design from the training phase, and those that do primarily focus on optimizing inference efficiency while overlooking training efficiency~\cite{sun2024you}. Since efficiency in both training and inference presents diverse design challenges, there is a growing need for a simple, holistic solution that addresses both aspects. 

In another direction, efforts for efficient training has presented methods for progressive growth~\cite{gong2019efficient,wu2024llama,Yano2025progressive}, progressive layer drop~\cite{zhang2020accelerating}, and progressive dataset complexity~\cite{li2022automated}. Conversely, efficient inference approaches have focused on model pruning~\cite{cheng2024survey}, distillation~\cite{yang2024survey}, progressive low-rank decomposition~\cite{hajimolahoseini2024single} and step-by-step distillation~\cite{hu2022lora,liu2024dora,hsieh2023distilling}. From a broader perspective, the progressive modification of model during training has proven superior to directly training the modified architecture, often with an additional advantage of many-in-one models~\cite{cai2024flextron}.

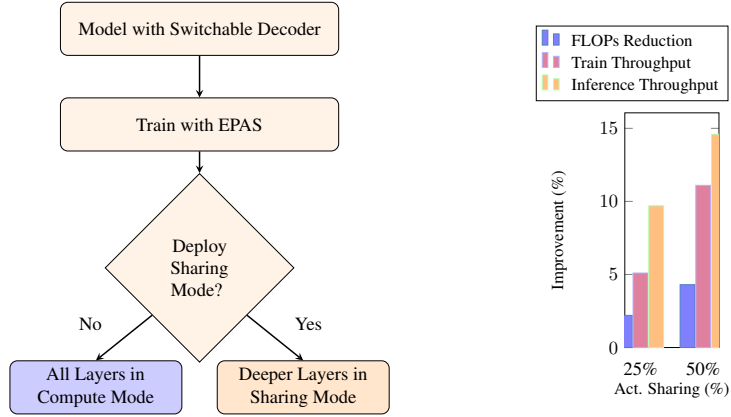
\begin{figure}
\resizebox{0.70\columnwidth}{!}{

\begin{tikzpicture}
\tikzset{
  box/.style={rounded corners, draw=#1, fill=#1!15, line width=0.7pt,
              minimum height=6mm, minimum width=22mm, align=center, font=\scriptsize},
  small/.style={rounded corners, draw=#1, fill=#1!15, line width=0.7pt,
                minimum height=6mm, minimum width=16mm, align=center, font=\scriptsize},
  add/.style ={circle, draw=gray!70, fill=gray!10, inner sep=0pt,
               minimum size=3.8mm, font=\scriptsize\bfseries},
  flow/.style={->, line width=0.7pt, line cap=round},
  labeltiny/.style={font=\scriptsize},
  label/.style={}
}
\colorlet{cNorm}{Goldenrod}     
\colorlet{cAttn}{ForestGreen}   
\colorlet{cProjL}{BurntOrange}       
\colorlet{cProjR}{NavyBlue}  
\colorlet{cDash}{BurntOrange}   

    \node (start) [process,rounded corners,fill=orange!10] at (0,6) {
    \parbox{5cm}{\centering{Model with Switchable Decoder}}
    };
    \node (process2) [process, rounded corners,fill=orange!10] at (0,4.2) {
    \parbox{5cm}{\centering{Train with EPAS }}
    };
    \node (decision) [decision,fill=orange!10] at (0,1.5) {
    \parbox{2cm}{
    \centering{
    Deploy\\
    Sharing Mode?
    }
    }
    };
    \node (output1) [process,rounded corners,fill=blue!20] at (-1.95,-0.75) {
     \parbox{3cm}{\centering{All Layers in\\ Compute Mode}}
    };
    \node (output2) [process, rounded corners,fill=orange!20] at (1.95, -0.75) {
         \parbox{3cm}{\centering{ Deeper Layers in\\ Sharing Mode}}
    };

\node[label] at ($(output1)+(-0.1,1.2)$) {No};
\node[label] at ($(output2)+(0.1,1.2)$) {Yes};
    
    \draw [arrow] (start) -- (process2);
    \draw [arrow] (process2) -- (decision);
    \draw [arrow] (decision.south west) -- (output1.north) node[midway, left]{}; 
    \draw [arrow] (decision.south east) -- (output2.north) node[midway, left]{};
        
    \begin{scope}[shift={(8,0)}] 
    \begin{axis}[
        width  = 0.24*\textwidth,
        height = 6cm,
        major x tick style = transparent,
        ybar=\pgflinewidth,
        bar width=8pt,
        ymajorgrids = false,
        ylabel = {\small{Improvement (\%)}},
        xlabel = {\small{Act. Sharing (\%)}},
        symbolic x coords={25\%,50\%},
        xtick = data,
        scaled y ticks = false,
        enlarge x limits=0.25,
        ymin=0,
        legend cell align=left,
        legend style={
                at={(1.1,1.05)},
                anchor=south east,
                column sep=1ex
        }
    ]
        \addplot[style={bblue,fill=blue!50,mark=none}]
            coordinates {(25\%, 2.2) (50\%,4.3)};
            
        \addplot[style={blue!30,fill=purple!50,mark=none}]
            coordinates {(25\%, 5.1) (50\%,11.1)};

        \addplot[style={green!30,fill=orange!50,mark=none}]
            coordinates {(25\%, 9.7) (50\%,14.6)};

        \legend{\small{FLOPs Reduction}, \small{Train Throughput}, \small{Inference Throughput}}
    \end{axis}
    \end{scope}

\end{tikzpicture}
}
\label{fig:intro}
\caption{\textbf{Left:} Overall solution from efficient training to inference using EPAS. \textbf{Right:}  TinyLLaMA~\cite{tinyllama} model FLOPs reduction and train/inference throughput improvement expanding $QK$ activation sharing to 25\% and 50\% of the layers.}
\end{figure}

The proposed progressive activation sharing combines progressive training and efficient inference in a unified training method for activation sharing models as shown a high level abstraction in Figure~\ref{fig:intro}. This method allows for utilizing redundancy observed in deeper layers from early phases of training while preserving model accuracy. It improves pretraining throughput to reduce time to accuracy and derives a family of efficient models from a single end-to-end training process. Additionally, it enables flexible transformation of pretrained models into activation sharing models through a single, efficient continual pretraining process without requiring multiple rounds of knowledge distillation. Instead of sharing activations during inference of a pretrained model, an activation-sharing region is progressively expanded during pretraining or continual pretraining. This makes computation lighter as training progresses. This hot switching to activation sharing during training is achieved through a switchable decoder block that can conditionally reuse activation. The training algorithm uses a scheduler that toggles activation‑sharing at configured intervals, progressively expanding the sharing block to improve training efficiency.
The key contributions are: 

\begin{itemize}
    \item EPAS enables faster training and flexible efficient model configurations during inference.
    \item EPAS transforms pretrained models into efficient ones through continual pretraining, eliminating the need for knowledge distillation.
    
    \item The proposed method achieved superior training and inference throughput while maintaining accuracy on par with baseline.
    
\end{itemize}


\section{Progressive Activation Sharing}
\label{sec:approach}
The proposed \textbf{E}fficient training with \textbf{P}rogressive \textbf{A}ctivation \textbf{S}haring (EPAS) method builds transformer models using the proposed switchable activation sharing decoder layer as building block. This decoder layer is a simple yet elegant extension that incorporates conditional activation sharing into the standard decoder layer. The term activation sharing generally refers to reusing some shared $QK$ or $KV$ activations from an early layer. This help to reduce the computation of somewhat redundant activations in the deeper parts of the model. At a high level, the proposed progressive activation sharing involves gradually growing the number of layers using activation sharing and hence increasing the throughput. In the following, we discuss the details of the newly designed decoder layer and the progressive training algorithm.

\subsection{Switchable Activation Sharing Decoder}
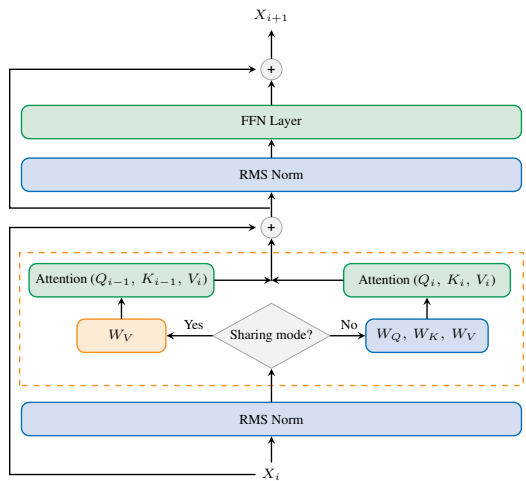
\begin{wrapfigure}{r}{0.5\columnwidth}
\resizebox{0.49\columnwidth}{!}{%
\begin{tikzpicture}[>=stealth]
\tikzset{
  box/.style={rounded corners, draw=#1, fill=#1!15, line width=0.7pt,
              minimum height=6mm, minimum width=22mm, align=center, font=\scriptsize},
  small/.style={rounded corners, draw=#1, fill=#1!15, line width=0.7pt,
                minimum height=6mm, minimum width=16mm, align=center, font=\scriptsize},
  add/.style ={circle, draw=gray!70, fill=gray!10, inner sep=0pt,
               minimum size=3.8mm, font=\scriptsize\bfseries},
  flow/.style={->, line width=0.7pt, line cap=round},
  labeltiny/.style={font=\scriptsize}
}
\colorlet{cNorm}{NavyBlue}     
\colorlet{cAttn}{ForestGreen}   
\colorlet{cProjL}{BurntOrange}       
\colorlet{cProjR}{NavyBlue}  
\colorlet{cDash}{BurntOrange}   

\node[labeltiny] (BL) at (0,0) {$X_i$};

\node[box=cNorm, minimum width=90mm] (RN1) at ($(BL)+(0,1.0)$) {RMS Norm};
\node[diamond, draw=gray!70, fill=gray!10, aspect=1.8, inner sep=1.2pt, font=\scriptsize] (DM) at ($(RN1)+(0,1.5)$) {Sharing mode?};

\node[small=cProjL] (WV) at ($(DM)+(-2.7,0)$) {$W_V$};
\node[small=cProjR, minimum width=22mm] (WQKV) at ($(DM)+(2.8,0)$) {$W_Q,~W_K,~W_V$};
\node[labeltiny] at ($(DM)+(-1.4,0.2)$) {Yes};
\node[labeltiny] at ($(DM)+(1.4,0.2)$) {No};

\node[box=cAttn, minimum width=30mm] (ATTNL)
  at ($(WV)+(0,1.0)$) {Attention ($Q_{i-1},\,K_{i-1},\,V_i$)};
\node[box=cAttn, minimum width=30mm] (ATTNR)
  at ($(WQKV)+(0,1.0)$) {Attention ($Q_i,\,K_i,\,V_i$)};
\draw[dashed, line width=0.7pt, cDash]
  ($(DM.south west)+(-4.0,-0.6)$) rectangle ($(DM.north east)+(4.0,1.2)$);
\coordinate (AfterATTN)  at ($(DM)+(0,1.0)$);

\node[add] (ADD1) at ($(AfterATTN)+(0,0.95)$) {+};
\coordinate (AfterADD1)  at ($(ADD1)+(0,0.25)$);

\node[box=cNorm, minimum width=90mm] (RN2) at ($(ADD1)+(0,0.95)$) {RMS Norm};

\node[box=cAttn, minimum width=90mm] (FFN) at ($(RN2)+(0,0.95)$) {FFN Layer};
  
\node[add] (ADD2) at ($(FFN)+(0,0.95)$) {+};

\node[labeltiny] (TL) at ($(ADD2)+(0,0.95)$) {$X_{i+1}$};

\draw[flow] (BL)  -- (RN1);
\draw[flow] (RN1) -- (DM);

\draw[flow] (DM) -- (WV);
\draw[flow] (DM) -- (WQKV);

\draw[flow] (WV) -- (ATTNL);
\draw[flow] (WQKV) -- (ATTNR);
\draw[flow] (ATTNL) -- (AfterATTN);
\draw[flow] (ATTNR) -- (AfterATTN);


\draw[flow] (AfterATTN) -- (ADD1);
\draw[flow] (ADD1) -- (RN2);
\draw[flow] (RN2)  -- (FFN);
\draw[flow] (FFN)  -- (ADD2);
\draw[flow] (ADD2) -- (TL);

\coordinate (L1) at ($(RN1.west)+(-0.2,0)$);
\coordinate (L2) at ($(RN2.west)+(-0.2,0)$);

\coordinate (ADD1_in)  at ($(ADD1.west)+(-0.8mm,0)$);  
\coordinate (ADD1_out) at ($(ADD1.west)+( 1.6mm,0.35)$);  
\coordinate (ADD2_in)  at ($(ADD2.west)+(-0.8mm,0)$);  

\draw[flow] (BL.west) -| (L1) |- (ADD1_in);

\draw[flow] (ADD1_out) -| (L2) |- (ADD2_in);

\end{tikzpicture}%
}
\caption{The Switchable Activation Sharing Decoder with an example of attention sharing ($Q,K$). This decoder layer extends conventional transformer decoder layer by adding a conditional switching branch to reuse $Q_{i-1},K_{i-1}$ from previous layer instead of computing in current layer(left branch inside \textcolor{BurntOrange}{dashed box}). When not using activation sharing mode, the computation follows the right branch as like convention decoder layer.}
\label{fig:epas_decoder}
\end{wrapfigure}

The hot switching of decoder layers to activation sharing mode during the progress of training is performed by switching a conditional branching of computation in the decoder layer. A pictorial illustration of this decoder extension, with a particular example of attention sharing, is presented in Figure~\ref{fig:epas_decoder}. This example demonstrates sharing of $Q,~K$ to make attention sharing compatible with Flash-Attention. The extension is quite simple without requiring any additional parameters. Hence, the modified architecture can reuse previously trained parameters for continual pretraining or post-training. This decoder simply branches out to either reuse some selected activations from a previous layer or to compute with current layer's own parameters. The most conventional activation sharing models use either attention scores, or $QK$, or $KV$ as the set of activations for this purpose. 

\subsection{Progressive Activation Sharing}
The progressive activation sharing approach aims to gradually expand the activation sharing region by switching decoder layers to activation sharing mode throughout the training steps. The growing of sharing layers follows a deterministic sharing strategy. The rationale behind deterministic sharing is to ensure that shared layers continuously grow, ultimately resulting in a model requiring fewer FLOPs than baseline during the inference.

\begin{figure*}[!t]
\centering
\begin{subfigure}[t]{.23\textwidth}
  \centering
  \begin{tikzpicture}[remember picture,baseline]
    \node (Abox) {\resizebox{\linewidth}{!}{%
      \begin{tikzpicture}
      \def\xsep{0.70cm}   
      \def\ysep{0.85cm}  
    
      \tikzset{
        neuron/.style={circle, draw=#1, fill=#1!18, line width=0.1pt, minimum size=1mm},
        layerlabel/.style={font=\scriptsize\bfseries, align=center}
      }
    
      \colorlet{c1}{NavyBlue}
      \colorlet{c2}{NavyBlue}
      \colorlet{c3}{NavyBlue}
      \colorlet{c4}{NavyBlue}
      \colorlet{c5}{NavyBlue}
    
      \coordinate (base) at (0,0);
    
      \begin{scope}[rotate=90]
        \newcommand{\threelayer}[4]{%
          \node[neuron=#3] (#2-2) at (#1) {};
          \node[neuron=#3] (#2-1) at ($(#1)+(0, 0.60*\ysep)$) {};
          \node[neuron=#3] (#2-3) at ($(#1)+(0,-0.60*\ysep)$) {};
          \node[layerlabel] at ($(#2-1)+(0,0.65*\ysep)$) {#4};
        }
    
        \coordinate (C1) at ($(base)+(0,0)$);
        \coordinate (C2) at ($(C1)+(\xsep,0)$);
        \coordinate (C3) at ($(C2)+(\xsep,0)$);
        \coordinate (C4) at ($(C3)+(\xsep,0)$);
        \coordinate (C5) at ($(C4)+(\xsep,0)$);
    
        \threelayer{C1}{L1}{c1}{}
        \threelayer{C2}{L2}{c2}{} 
        \threelayer{C3}{L3}{c3}{} 
        \threelayer{C4}{L4}{c4}{} 
        \threelayer{C5}{L5}{c5}{} 
    
        \foreach \a in {1,2,3} \foreach \b in {1,2,3} \draw[->,>=stealth, line width=0.15pt] (L1-\a) -- (L2-\b);
        \foreach \a in {1,2,3} \foreach \b in {1,2,3} \draw[->,>=stealth, line width=0.15pt] (L2-\a) -- (L3-\b);
        \foreach \a in {1,2,3} \foreach \b in {1,2,3} \draw[->,>=stealth, line width=0.15pt] (L3-\a) -- (L4-\b);
        \foreach \a in {1,2,3} \foreach \b in {1,2,3} \draw[->,>=stealth, line width=0.15pt] (L4-\a) -- (L5-\b);
      \end{scope}
    \end{tikzpicture}%
    }};
  \end{tikzpicture}
  \caption{Baseline}
\end{subfigure}\hfill
\begin{subfigure}[t]{.23\textwidth}
  \centering
  \begin{tikzpicture}[remember picture,baseline]
    \node (Bbox) {\resizebox{\linewidth}{!}{%
         \begin{tikzpicture}
      \def\xsep{0.70cm}   
      \def\ysep{0.85cm}  
    
      \tikzset{
        neuron/.style={circle, draw=#1, fill=#1!18, line width=0.1pt, minimum size=1mm},
        layerlabel/.style={font=\scriptsize\bfseries, align=center}
      }
    
      \colorlet{c1}{NavyBlue}
      \colorlet{c2}{NavyBlue}
      \colorlet{c3}{NavyBlue}
      \colorlet{c4}{NavyBlue}
      \colorlet{c5}{BurntOrange}
    
      \coordinate (base) at (0,0);
    
      \begin{scope}[rotate=90]
        \newcommand{\threelayer}[4]{%
          \node[neuron=#3] (#2-2) at (#1) {};
          \node[neuron=#3] (#2-1) at ($(#1)+(0, 0.60*\ysep)$) {};
          \node[neuron=#3] (#2-3) at ($(#1)+(0,-0.60*\ysep)$) {};
          \node[layerlabel] at ($(#2-1)+(0,0.65*\ysep)$) {#4};
        }
    
        \coordinate (C1) at ($(base)+(0,0)$);
        \coordinate (C2) at ($(C1)+(\xsep,0)$);
        \coordinate (C3) at ($(C2)+(\xsep,0)$);
        \coordinate (C4) at ($(C3)+(\xsep,0)$);
        \coordinate (C5) at ($(C4)+(\xsep,0)$);
    
        \threelayer{C1}{L1}{c1}{}
        \threelayer{C2}{L2}{c2}{} 
        \threelayer{C3}{L3}{c3}{} 
        \threelayer{C4}{L4}{c4}{} 
        \threelayer{C5}{L5}{c5}{} 
    
        \foreach \a in {1,2,3} \foreach \b in {1,2,3} \draw[->,>=stealth, line width=0.15pt] (L1-\a) -- (L2-\b);
        \foreach \a in {1,2,3} \foreach \b in {1,2,3} \draw[->,>=stealth, line width=0.15pt] (L2-\a) -- (L3-\b);
        \foreach \a in {1,2,3} \foreach \b in {1,2,3} \draw[->,>=stealth, line width=0.15pt] (L3-\a) -- (L4-\b);
        \foreach \a in {1,2,3} \foreach \b in {1,2,3} \draw[->,>=stealth, line width=0.15pt] (L4-\a) -- (L5-\b);
      \end{scope}
    \end{tikzpicture}%
    }};
  \end{tikzpicture}
  \caption{Begin activation sharing}
\end{subfigure}\hfill
\begin{subfigure}[t]{.23\textwidth}
  \centering
  \begin{tikzpicture}[remember picture,baseline]
    \node (Cbox) {\resizebox{\linewidth}{!}{%
            \begin{tikzpicture}
      \def\xsep{0.70cm}   
      \def\ysep{0.85cm}  
    
      \tikzset{
        neuron/.style={circle, draw=#1, fill=#1!18, line width=0.1pt, minimum size=1mm},
        layerlabel/.style={font=\scriptsize\bfseries, align=center}
      }
    
      \colorlet{c1}{NavyBlue}
      \colorlet{c2}{NavyBlue}
      \colorlet{c3}{NavyBlue}
      \colorlet{c4}{BurntOrange}
      \colorlet{c5}{BurntOrange}
    
      \coordinate (base) at (0,0);
    
      \begin{scope}[rotate=90]
        \newcommand{\threelayer}[4]{%
          \node[neuron=#3] (#2-2) at (#1) {};
          \node[neuron=#3] (#2-1) at ($(#1)+(0, 0.60*\ysep)$) {};
          \node[neuron=#3] (#2-3) at ($(#1)+(0,-0.60*\ysep)$) {};
          \node[layerlabel] at ($(#2-1)+(0,0.65*\ysep)$) {#4};
        }
    
        \coordinate (C1) at ($(base)+(0,0)$);
        \coordinate (C2) at ($(C1)+(\xsep,0)$);
        \coordinate (C3) at ($(C2)+(\xsep,0)$);
        \coordinate (C4) at ($(C3)+(\xsep,0)$);
        \coordinate (C5) at ($(C4)+(\xsep,0)$);
    
        \threelayer{C1}{L1}{c1}{}
        \threelayer{C2}{L2}{c2}{} 
        \threelayer{C3}{L3}{c3}{} 
        \threelayer{C4}{L4}{c4}{} 
        \threelayer{C5}{L5}{c5}{} 
    
        \foreach \a in {1,2,3} \foreach \b in {1,2,3} \draw[->,>=stealth, line width=0.15pt] (L1-\a) -- (L2-\b);
        \foreach \a in {1,2,3} \foreach \b in {1,2,3} \draw[->,>=stealth, line width=0.15pt] (L2-\a) -- (L3-\b);
        \foreach \a in {1,2,3} \foreach \b in {1,2,3} \draw[->,>=stealth, line width=0.15pt] (L3-\a) -- (L4-\b);
        \foreach \a in {1,2,3} \foreach \b in {1,2,3} \draw[->,>=stealth, line width=0.15pt] (L4-\a) -- (L5-\b);
      \end{scope}
    \end{tikzpicture}%
    }};
  \end{tikzpicture}
  \caption{Grow sharing region}
\end{subfigure}\hfill
\begin{subfigure}[t]{.23\textwidth}
  \centering
  \begin{tikzpicture}[remember picture,baseline]
    \node (Dbox) {\resizebox{\linewidth}{!}{%
            \begin{tikzpicture}
      \def\xsep{0.70cm}   
      \def\ysep{0.85cm}  
    
      \tikzset{
        neuron/.style={circle, draw=#1, fill=#1!18, line width=0.1pt, minimum size=1mm},
        layerlabel/.style={font=\scriptsize\bfseries, align=center}
      }
    
      \colorlet{c1}{NavyBlue}
      \colorlet{c2}{NavyBlue}
      \colorlet{c3}{BurntOrange}
      \colorlet{c4}{BurntOrange}
      \colorlet{c5}{BurntOrange}
    
      \coordinate (base) at (0,0);
    
      \begin{scope}[rotate=90]
        \newcommand{\threelayer}[4]{%
          \node[neuron=#3] (#2-2) at (#1) {};
          \node[neuron=#3] (#2-1) at ($(#1)+(0, 0.60*\ysep)$) {};
          \node[neuron=#3] (#2-3) at ($(#1)+(0,-0.60*\ysep)$) {};
          \node[layerlabel] at ($(#2-1)+(0,0.65*\ysep)$) {#4};
        }
    
        \coordinate (C1) at ($(base)+(0,0)$);
        \coordinate (C2) at ($(C1)+(\xsep,0)$);
        \coordinate (C3) at ($(C2)+(\xsep,0)$);
        \coordinate (C4) at ($(C3)+(\xsep,0)$);
        \coordinate (C5) at ($(C4)+(\xsep,0)$);
    
        \threelayer{C1}{L1}{c1}{}
        \threelayer{C2}{L2}{c2}{} 
        \threelayer{C3}{L3}{c3}{} 
        \threelayer{C4}{L4}{c4}{} 
        \threelayer{C5}{L5}{c5}{} 
    
        \foreach \a in {1,2,3} \foreach \b in {1,2,3} \draw[->,>=stealth, line width=0.15pt] (L1-\a) -- (L2-\b);
        \foreach \a in {1,2,3} \foreach \b in {1,2,3} \draw[->,>=stealth, line width=0.15pt] (L2-\a) -- (L3-\b);
        \foreach \a in {1,2,3} \foreach \b in {1,2,3} \draw[->,>=stealth, line width=0.15pt] (L3-\a) -- (L4-\b);
        \foreach \a in {1,2,3} \foreach \b in {1,2,3} \draw[->,>=stealth, line width=0.15pt] (L4-\a) -- (L5-\b);
      \end{scope}
    \end{tikzpicture}%
    }};
  \end{tikzpicture}
  \caption{Activation sharing model}
\end{subfigure}

\begin{tikzpicture}[remember picture,overlay]
  \tikzset{flow/.style={-{Latex[length=6mm,width=3.5mm]},
                        line width=1.4pt, line cap=round,
                        shorten >=1.2mm, shorten <=1.2mm}}

  \draw[flow] ([xshift=6mm]Abox.east) -- ([xshift=5mm]Bbox.west);
  \draw[flow] ([xshift=6mm]Bbox.east) -- ([xshift=5mm]Cbox.west);
  \draw[flow] ([xshift=6mm]Cbox.east) -- ([xshift=5mm]Dbox.west);
\end{tikzpicture}

\caption{An example of EPAS training approach. Beginning with all $L$ layers in \textcolor{NavyBlue}{\textbf{compute-mode}} (e.g., 5 layers here), a region of $B$ layers transition into \textcolor{BurntOrange}{\textbf{sharing-mode}} ( e.g., 1 layer here) at every $I$ step intervals. The progressive growth continues till maximum $S$ layers (e.g., 3 layers here) are in sharing mode. The trained model can be used either all layer in compute mode or up to $S$ layers in activation sharing mode.}
\label{fig:epas_steps}
\end{figure*}
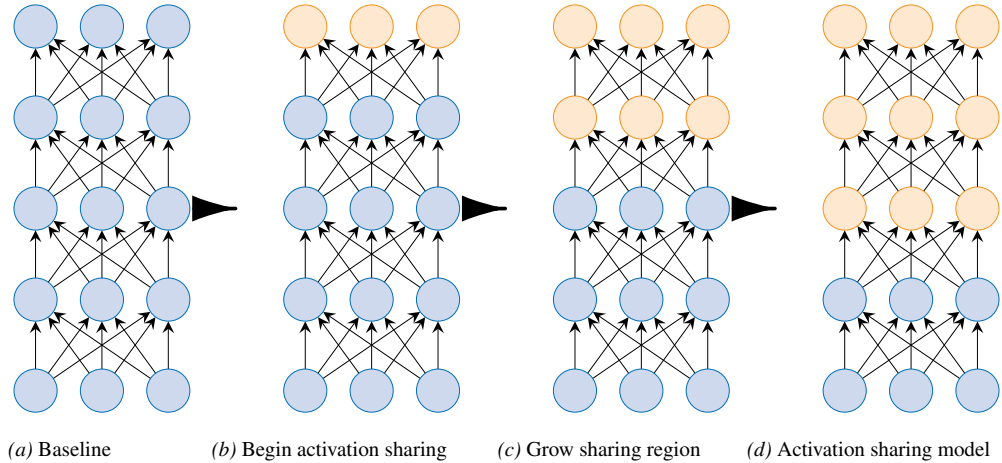

Initially, training begins with all of the $L$ layers of a model, $M$, in compute mode. At this stage the model works like a conventional transformer model~\cite{vaswani2017attention}. A target sharing region, $S$, defines a list of layers that will progressively transition into activation sharing mode. We found that selecting a set of deeper layers and activated sharing sequentially from deep to shallow layers works well in this scenario. Over the course of $T$ training steps, a group of $B$ layers at the deep end of $S$ is switched to activation sharing mode at every interval of $I$ training steps. Rather than enabling activation sharing of the whole sharing region from the beginning of training, the method gradually expands the sharing region by gently allowing the model to adapt to activation sharing. Following this training strategy, it ends up with a target model of a predefined maximum activation sharing region. The steps are depicted with a schematic example in Figure~\ref{fig:epas_steps}.

In this proposed activation sharing scheme, the layer immediately before the sharing region shares its activations with the layers in sharing group. During the forward pass, if a layer detects that its subsequent layer is in sharing mode, it populates an activation cache with a selected set of activations, $A$. This progressive activation sharing method is compatible with sharing attention scores, $QK$, or $KV$. The trained model benefits from reduced model FLOPs by leveraging the last state of the activation sharing group. Algorithm~\ref{alg:epass_algo} presents the progressive activation sharing training algorithm instantiated with $QK$ sharing; the adaptation to other forms of activation sharing, such as $KV$ sharing is straightforward.


\begin{algorithm}[t]
\resizebox{0.85\columnwidth}{!}{%
\begin{minipage}{0.84\columnwidth}
\caption{Progressive Activation Sharing }
\label{alg:epass_algo}
\begin{algorithmic}[1]
\State \textbf{Input:} Model, $M$, Interval, $I$, Target Sharing Layers, $S_c$, Sharing Region Growth Size $B$
\State \textbf{Output:} Trained model $M$
\State \textbf{$L$:} Layers in Base Model
\State \textbf{$T$:} Training Steps
\State \textbf{$S$:} Layers currently in sharing mode

\State \textbf{assert} $(B \geq 1 \And |S_c| \geq 1)$
\State \textbf{assert} $ (|S_c| \mod B = 0)$

\State $C \gets [] $, $S \gets [] $ 

\For{$t = 1$ to $T$} 
    \For{$i = 0$ to $|L|-1$} 
        \If{ $L[i] \in S$}
             \State $Q_i, K_i \gets Q_{i-1},~K_{i-1}$ from $C$ 
             \State Compute $V_i$
        \Else 
             \State Compute $Q_i,~K_i,V_i$
        \EndIf
        \State Do the rest of the computations 
        \If{ $L[i+1]$ has sharing on} 
             \State $C \gets $ $Q_i,K_i$ 
        \EndIf
    \EndFor
    \If {$ t \% I == 0 \And |S_c| \geq B$}
        \State $L_c \gets $ pop last  $B$ element from $S_c$ 
        \State $S \gets L_c + S$  \Comment{ Append beginning}
    \EndIf
\EndFor
\State return $M$
\end{algorithmic}
\end{minipage}
}
\end{algorithm}

\subsection{Applications}


Progressively growing the sharing block during training gradually reduces model FLOPs and increases training throughput (tokens/sec). This results in reduced training time and cost, while the gradual change in the computation graph allows for training stability. EPAS enhances computational resource utilization and improves pretraining efficiency by minimizing redundant activations, thereby achieving a balanced trade-off between efficiency and model performance. EPAS can also transform existing pretrained models into activation-sharing architectures while used in continual pretraining setting. 
Moreover, continual pretraining with EPAS enables flexible sub-network selection during inference, allowing a single end to end training on a small dataset to derive a family of efficient models. This eliminates the need for complex multi-model training or repeated distillation. 

During the inference phase, the activation sharing models trained with EPAS can achieve superior throughput compared to baseline models while maintaining similar performance. Attention score sharing, or $QK$ sharing, reduces computational overhead and minimizes $KV$ cache memory requirement as the sharing layers only need to cache $V$. Conversely, $KV$ sharing saves more memory as it eliminates the need to cache both K and V in the sharing layers while offering less computational reduction. In either of the case, the inference process becomes faster and more resource-efficient compared to baseline model. 



\section{Experiments}
\label{sec:experiements}
To demonstrate the efficacy of EPAS, we considered $QK$ sharing as a particular instance of activation sharing. The experiments integrate EPAS with open-source transformer-based LLM. In particular, we perform extensive experiment with TinyLLaMA-1.1B~\cite{tinyllama} as our primary baseline and then further extend our empirical analysis with more LlaMA-based models~\cite{touvron2023llama,touvron2023llama2} ranging from 125M to 7B parameters. We used a small subset of the open-source SlimPajama-627B dataset for the pretraining and continual pretraining experiments~\cite{SlimPajama627B}. Since our pretraining experiment is focused on efficiency analysis and comparing training dynamics for small number of training steps rather than training till convergence, this set up is sufficient for the intended purpose. 

The empirical analysis compared training and inference efficiency as well as learning capacity during training. We measure theoretical FLOPs reduction, training throughput (tokens/sec) and inference throughput (tokens/sec). For evaluation of transformed pretrained model with continual pretraining setup of EPAS, we used lm-eval-harness~\cite{eval-harness}. Furthermore, to compare the efficiency across diverse devices, experiments are conducted on Nvidia V100 GPU, Ascend 910A NPU, and Ascend 910B NPU.


Furthermore, we present extensive ablation study for critical understanding and justification of our findings and corresponding design choices. While the proposed method is compatible to share various activations, for the scope of this research we limit empirical analysis on attention sharing only. In particular, we used cross-layer query and key ($QK$) sharing so that the method is compatible with Flash-Attention~\cite{dao2022flashattention,dao2023flashattention}.

\subsection{Training efficiency}
To empirically assess training efficiency, we conduct model FLOPs and training throughput analysis by scaling model sizes from 125M to 7B following LLaMA architectures. Although EPAS presents a generalized training algorithm to train activation sharing architecture, we demonstrate for the example of half of the layers in final sharing mode following recent trends~\cite{sun2024you}. Table~\ref{tab:model_flops} summarizes the reduction of theoretical model FLOPs for each of the model when sharing $QK$ across second half of the layers. We observe up to 8\% FLOPs reduction with this configuration. Table~\ref{tab:pretraining_efficiency_scaling} presents the improvement in training tokens/second with a distributed training setup on 8 V100 GPU. The models show up to 11\% train throughput improvement with the above configuration. 

\begin{table}[ht]
\centering
\begin{minipage}{0.45\textwidth}
\centering
\begin{tabular}{l|ccc}
\hline
\multirow{2}{*}{Model} & \multicolumn{3}{c}{Model FLOPs/sample (TF)}\\
\cline{2-4}
& Baseline & Q/K-Sharing & Reduction(\%) \\
\hline
125M&  1.81 & 1.72 & {4.9\%}  \\
\hline 
1.1B & 14.98 & 14.34 & {4.3\%} \\  
\hline 
3B & 41.41 & 38.39 & {7.3\%} \\
\hline 
7B & 85.62 & 79.21 & {8.1\%} \\ 

\hline
\end{tabular}
\caption{Model FLOPs per sample in Terra-FLOPs (TF) for baseline versus $QK$ sharing of 50\% of the layers.} 
\label{tab:model_flops}

\end{minipage}
\hfill
\begin{minipage}{0.45\textwidth}
\begin{tabular}{l|ccc}
\hline
\multirow{2}{*}{Model} & \multicolumn{3}{c}{Train Tokens/sec}\\
\cline{2-4}
& Baseline & Q/K-Sharing & Improvement(\%) \\
\hline
125M&  15458.9 & 17143.1 & {10.8\%}  \\
\hline 
1.1B & 3850.2 & 4259.8&  {11.1\%} \\ 
\hline 
3B & 1783.1 & 1961.3 & {10.9\%}  \\ 
\hline 
7B & 1259.8 & 1367.9 & {8.6\%}  \\ 
\hline
\end{tabular}
\caption{Model size scaling and training efficiency of $QK$ sharing on V100 GPU with distributed training setup. }
\label{tab:pretraining_efficiency_scaling}
\end{minipage}
\end{table}

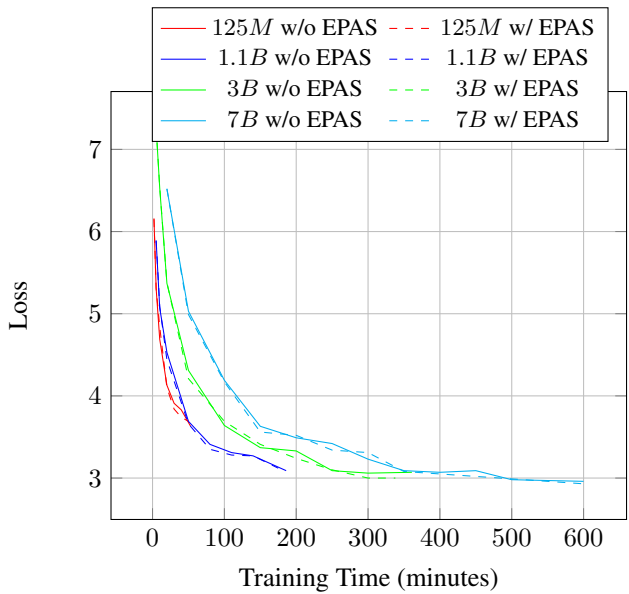
\begin{wrapfigure}{r}{0.58\columnwidth}
\begin{tikzpicture}
    \begin{axis}[
        xlabel={Training Time (minutes)},
        ylabel={Loss},
        grid=major,
        legend columns=2,
        legend style={ at={(axis cs:-1.0,8.73)},
                /tikz/column 2/.style={
                column sep=2pt},
                anchor=north west}, 
        ]
        \addplot[color=red, solid] coordinates {
            (2,6.16)(5,5.31)(10,4.69)(20,4.13)(30,3.91)(40,3.83)(53,3.63)
        };
        \addlegendentry{\small{$125M$ w/o EPAS}}

        \addplot[color=red, dashed] coordinates {
            (2,6.15)(5,5.31)(10,4.84)(20,4.10)(30,3.84)(40,3.74)(50,3.69)
        };
        \addlegendentry{\small{$125M$ w/ EPAS}}

        \addplot[color=blue, solid] coordinates {
            (5,5.89)(10,5.07)(20,4.53)(50,3.69)(80,3.41)(110,3.31)(140,3.27)(186,3.09)
        };
        \addlegendentry{\small{$1.1B$ w/o EPAS}}

        \addplot[color=blue, dashed] coordinates {
            (5,5.89)(10,5.09)(20,4.44)(50,3.66)(80,3.35)(110,3.28)(140,3.27)(175,3.12)
        };
        \addlegendentry{\small{$1.1B$ w/ EPAS}}

        \addplot[color=green, solid] coordinates {
            (5,7.27)(10,6.53)(20,5.37)(50,4.31)(100,3.64)(150,3.37)(200,3.33)(250,3.09)(300,3.06)(361,3.07)
        };
        \addlegendentry{\small{$3B$ w/o EPAS}}

        \addplot[color=green, dashed] coordinates {
            (5,7.27)(10,6.54)(20,5.38)(50,4.21)(100,3.69)(150,3.41)(200,3.24)(250,3.10)(300,3.00)(338,3.00)
        };
        \addlegendentry{\small{$3B$ w/ EPAS}}

        \addplot[color=cyan, solid] coordinates {
            (20,6.52)(50,5.03) (100,4.19) (150,3.63) (200,3.49) (250,3.42 ) (300,3.23) (350,3.09) (400,3.07)(450,3.09)(500,2.98)(600,2.96)
        };
        \addlegendentry{\small{$7B$ w/o EPAS}}

        \addplot[color=cyan, dashed] coordinates {
            (20,6.52)(50,4.99) (100,4.17) (150,3.56) (200,3.52) (250,3.34) (300,3.31) (350,3.08) (400,3.05)(500,2.99)(600,2.93)
        };
        \addlegendentry{\small{$7B$ w/ EPAS}}

    \end{axis}
\end{tikzpicture}
\caption{Smoothed loss versus time while scaling model sizes across LLaMA models with $125M$, $1.1B$, and $3B$ parameters. $QK$ sharing models trained with EPAS also exhibit slightly faster convergence during training in addition to have higher throughput during inference.} 
\label{fig:pt_loss_vs_time_llama1B}
\end{wrapfigure}


\begin{table*}
\centering
\begin{adjustbox}{width=0.99\textwidth}
\begin{tabular}{ll|ccc|ccc}
\hline
\multirow{2}{*}{Device Name}  & \multirow{2}{*}{Device Spec} &  \multicolumn{3}{c|}{Single Device} & \multicolumn{3}{c}{Distributed (8 Device)} \\
\cline{3-8}
&  & Baseline & Q/K Sharing & Improvement(\%) & Baseline & Q/K Sharing & Improvement(\%)\\
\hline 
V100 GPU & 32GB, 125 TF & 4079.6  & 4423.6& {10.8\%} & 3850.2 & 4259.8&  {11.1\%}\\
910A NPU & 32GB, 278 TF &  4321.3 &4874.2& {12.8\%}&4788.2&5246.9 & {9.57\%} \\
910B NPU & 64GB, 378 TF &  11354.1 & 12443.7 &  {9.6\%} &12902.1&13844.5  & {7.3\%} \\
\hline
\end{tabular}
\end{adjustbox}
\caption{Empirical evidence of training efficiency comparing throughput (tokens/sec) across different hardware for activation sharing ($QK$) in the second half of the layers of the TinyLLaMA model.
}
\label{tab:pretraining_efficiency}
\end{table*}

We further investigated the learning capacity of activation sharing model when trained with EPAS compared to training baseline model without activation sharing. This experiment was conducted for $0.25M$ tokens per step for $4000$ steps resulting in a total of $1B$ tokens. We consider this setup sufficient for comparing the training dynamics of the baseline and activation sharing models by analyzing loss curves without conducting a full LM benchmark evaluation similar to recent literature~\cite{sun2024you}. 

We observe that EPAS has lower loss at equal time and needs less time to achieve equal loss as baseline. We present a loss curve comparison for scaling analysis of training w/ and w/o EPAS in Figure~\ref{fig:pt_loss_vs_time_llama1B}. The comparison of train loss vs. time shows faster training and convergence with EPAS while maintaining similar loss curve pattern as the baseline. This is particularly evident from the observation that at any given time during training, EPAS shows a lower loss, especially in the early phases of training.

The results in Table~\ref{tab:val_loss} compare the total training time and the final validation loss after training. The findings show that EPAS significantly speeds up training, while the difference in final validation loss remains negligibly small (less than 0.05), indicating faster convergence without sacrificing accuracy or increasing the risk of overfitting.

\begin{wraptable}{r}{0.65\textwidth}
\centering
\begin{tabular}{l|cc}
\hline
\textbf{Model} &\textbf{Train Time (hh:mm:ss)} & \textbf{Validation Loss} \\
\hline 
125M (w/o EPAS) & 00:54:12 & 3.74 \\    
\hline
125M (w/ EPAS) & 00:50:50 & 3.79 \\    
\hline
\hline 
1.1B (w/o EPAS) & 03:07:18 & 3.19 \\    
\hline
1.1B (w/ EPAS) & 02:56:33 & 3.22 \\    
\hline
\hline 
3B (w/o EPAS) & 06:02:46 & 3.06 \\    
\hline
3B (w/ EPAS) & 05:39:11 & 3.09 \\    
\hline
\hline 
7B (w/o EPAS) & 11:09:26 & 2.99 \\    
\hline
7B (w/ EPAS) & 10:25:17 & 3.04 \\    
\hline
\end{tabular}
\caption{Total time and final validation loss for several LLaMA models of varying parameter sizes, demonstrating that EPAS training is significantly faster with a negligible difference in final validation loss.}.
\label{tab:val_loss}
\end{wraptable}

We also considered extending our experiments for various hardware types. For this cross hardware experiment, we fix a model (TinyLLaMA) and perform the same analysis across various hardware. Table~\ref{tab:pretraining_efficiency} presents the observation of training efficiency across different hardware categories. The table shows a negligibly minor variation in train throughput improvement while changing hardware. In particular, the $QK$ sharing model still maintains 8-10\% train throughput improvement across the three types of hardware. This evidence further justifies that activation sharing makes a generic efficiency improvement on model's computation that persists across varieties of device specs.

\begin{wraptable}{r}{0.50\textwidth}
\centering
\begin{adjustbox}{width=0.49\textwidth}
\begin{tabular}{l|ccc}
\hline
\multirow{2}{*}{Model} & \multicolumn{3}{c}{Inference Throughput(tok/sec)}\\
\cline{2-4}
& Baseline & 25\% Sharing & 50\% Sharing \\
\hline
125M &  63.8 & 78.3 ({22.7\%}) & 82.4 ({29.2\%})  \\
\hline
1.1B &  39.1 & 42.9 ({9.7\%}) & 44.8 ({14.6\%})  \\
\hline 
3B & 38.1 & 40.9 ({7.3\%}) & 43.2 ({13.4\%})\\  
\hline 
7B & 24.46 & 25.27 ({3.3\%}) & 26.03 ({6.4\%}) \\ 
\hline
\end{tabular}
\end{adjustbox}
\caption{Inference throughput on V100 GPU with $QK$ sharing on 25\% and 50\% of the layer in sharing mode.}
\label{tab:gen_throughput}
\end{wraptable}

\subsection{Inference efficiency }
We compare the generation throughput in terms of tokens/sec to measure the inference efficiency improvement of the proposed method. Since models trained with EPAS present a flexible architecture that can be used with various number of activation sharing layers during inference, we present inference throughput for sharing $QK$ activations for 25\% and 50\% of the layers. For example, LLaMA1.1B has 22 layers, hence 25\% and 50\% indicates 5 and 11 layers in sharing mode respectively. We observe a 3–22\% improvement in generation throughput when sharing 25\% of the layers, and a 6–30\% improvement when sharing 50\% of the layers across different models. The results are summarized Table~\ref{tab:gen_throughput}.

\begin{table*}[!ht]
\centering
\begin{adjustbox}{width=0.99\textwidth}
\begin{tabular}{l|ccccccccccc|c}
\hline
Model  & WG & PIQA & BoolQ & ARC-C & ARC-E & OBQA & HS & SciQ & LM(oa) & LM(std) & RTE &Average \\
\hline 
w/o EPAS (No Sharing) & 59.12& 73.56 & 56.09 & 32.68 & 55.51& 36.80 & 61.45& 84.20 & 56.70 & 50.55& 57.04 & 56.70 \\
w/ EPAS (No Sharing)        & 58.25 & 73.01 & 63.33 & 31.57 & 56.44 & 36.20 & 58.17 & 85.90 & 56.10 & 52.45 & 56.32 & 57.07 \\
\hline \hline
w/o EPAS (Sharing 3/22)$\dagger$ & 60.14 & 73.50 & 48.41 & 32.25 & 53.91 & 37.40 & 60.48 & 75.80 & 25.00 & 23.17 & 47.65 & 48.88\\ 
w/ EPAS (Sharing 3/22) & 59.27 & 73.07 & 62.72 & 33.02 & 56.14 & 36.20 & 58.72 & 83.60 & 49.97 & 47.80 & 52.35 & 55.71 \\ 
\hline \hline
w/o EPAS (Sharing 5/22)$\dagger$ & 55.64 & 67.03 & 48.41 & 27.13 & 42.55 & 31.00 & 51.51 & 75.80 & 25.00 & 23.17 & 47.65 & 44.99 \\
w/ EPAS (Sharing 5/22) & 58.02 & 73.18 & 60.70 & 31.23 & 55.85 & 36.00 & 57.95 & 83.10 & 47.58 & 44.69 & 50.90 & 54.47 \\
\hline 
\end{tabular}
\end{adjustbox}
\caption{Evaluation of multiple inference configurations from a single trained model. The EPAS model uses continual pretraining with a target sharing region of five layers while varied sharing region length during inference.  $\dagger$ represents our re-implementation of ~\cite{liao2024beyond} for smaller model for various $QK$ sharing settings.}
\label{tab:cpt_eval_results}
\end{table*}

\subsection{Language Model Evaluations}

We consider to evaluate model trained with EPAS in a continual pretraining setup to transform base models to attention-sharing models. We follow this direction due to the resource intensity of pretraining from scratch. For fair comparison, we also do continual pretraining on base models with same dateset. This experiment begins with a pretrained checkpoint and follows continual pretraining with progressively growing the activation sharing block from the deep end of the model towards the shallow end gradually growing one large activation sharing region. No additional sophisticated training approaches, e.g., knowledge distillation or additional auxiliary loss computation  are used. In particular, we consider a pretrained TinyLLaMA model as baseline. Due to limited resources we constrain the activation sharing block to grow up to 25\% of the model depth( e.g., 5 out of 22 layers). The training is done in a distributed setup of 8 device with a batch size 8 per device and gradient accumulation steps 16 to match the effective token per step being equal to the baseline pretraining configuration ( e.g., $8\times 8 \times 16 \times 2048 = 2 M$). The training is run for $2K$ steps resulting in total $4$ billion tokens. We use the lm-eval-harness~\cite{eval-harness} for evaluating the models on LM benchmarks.

The checkpoint from continual pretraining with EPAS is evaluated against applying inference time attention sharing, known as Beyond-KV-Cache~\cite{liao2024beyond} under various activation sharing configurations. We first compare the baseline and EPAS trained model with both of the models in full-compute mode. Then, we compare by using three and five layers in sharing mode. We observe that the model trained without EPAS shows a large drop in accuracy when evaluated with sharing mode. In contrast, EPAS trained model can retain most of the accuracy in attention sharing mode. When using five layers in sharing mode for both of the models, the EPAS trained model shows around 10\% higher accuracy. The results are presented in Table~\ref{tab:cpt_eval_results}, where each row pair shares the same activation sharing configuration.

Full model evaluation during inference without activation sharing improves accuracy in the EPAS-trained model. As expected, baseline accuracy declines more rapidly as additional layers are included in the sharing block. In contrast, the EPAS-trained model maintains robust accuracy as the activation sharing block expands. Notably, with a $QK$ sharing block size of five layers ( 25\% of model depth), the EPAS-trained model outperforms the baseline by approximately 12\%. These findings highlight EPAS’s effectiveness in preserving accuracy while progressively sharing activations, offering a robust approach for efficient transformer model training and inference.



\begin{wraptable}{r}{0.55\textwidth}
\centering
\begin{adjustbox}{width=0.54\columnwidth}
\begin{tabular}{l|ccccc}
\hline
\textbf{Model} &\textbf{PIQA} & \textbf{WG} &\textbf{BoolQ} & \textbf{OBQA} & \textbf{HS}\\
\hline
Excluding last layer& 73.29 & 58.01 & 52.94 & 36.00 & 58.66 \\
\hline 
Including last layer & \textbf{73.39} & \textbf{59.12} & \textbf{56.51} & \textbf{36.80} & \textbf{59.11} \\    
\hline
\end{tabular}
\end{adjustbox}
\caption{LM benchmark evaluation of continual pretraining reveals a minor difference between including or excluding the last layer in activation sharing.
}
\label{tab:ab_last_layer}
\end{wraptable}

\subsection{Ablation Experiment}

\textbf{Impact of Last Layer.} Recent research presents diverging perspectives on activation sharing regarding the role of the final layer. One approach argues for excluding the last layer or retaining it solely during inference due to its distinct attention pattern~\cite{liao2024beyond}, while an alternative view supports its inclusion within the activation-sharing block~\cite{sun2024you}. To assess the optimal configuration, we conducted an analysis using continual pretraining of TinyLLaMA with EPAS, evaluating LM benchmark performance on a small dataset under two conditions: exclusion versus inclusion of the last layer within sharing blocks. As summarized in Table~\ref{tab:ab_last_layer}, the results indicate minimal performance differences, with a slight advantage observed when incorporating the last layer during training. Consequently, we adopt this strategy in our approach.

\begin{figure*} 
\begin{center}    
\resizebox{0.70\columnwidth}{!}{
\begin{tikzpicture}
    \foreach \i in {1,2,3,4,5,6} {
        \ifodd\i
            \def\colorbox{blue!20}
        \else
            \def\colorbox{orange!20}
        \fi

        \node[draw, rounded corners, fill=\colorbox, minimum width=2cm, minimum height=1cm] (rect\i) at (\i*2.5, 4) {$L_{i+\i}$};
        \ifnum\i>1
            \draw[->] (rect\the\numexpr\i-1\relax.east) -- (rect\i.west);
        \fi
    }

    \foreach \i in {2,4,6} {
        \node[draw, circle, fill=yellow!50, minimum size=0.6cm] (circle\i) at ({(\i-1)*2.5+1.25}, 5.5) {};
        \draw[->] (rect\the\numexpr\i-1\relax.north) -- (circle\i.west);
        \draw[->] (circle\i.east) -- (rect\i.north);
    }
    \node at (8, 3.1) {\textbf{(a)} Sharing activation in multiple small regions of alternate layers.};
    
    \foreach \i in {1,2,3} {
        \node[draw, rounded corners, fill=blue!30, minimum width=2cm, minimum height=1cm] (rectB\i) at (\i*2.5, 1) {$L_{i+\i}$};
        \ifnum\i>1
            \draw[->] (rectB\the\numexpr\i-1\relax.east) -- (rectB\i.west);
        \fi
    }
    
    \foreach \i in {4,5,6} {
        \node[draw, rounded corners, fill=orange!30, minimum width=2cm, minimum height=1cm] (rectB\i) at (\i*2.5, 1) {$L_{i+\i}$};
        \draw[->] (rectB\the\numexpr\i-1\relax.east) -- (rectB\i.west);
    }

    \node[draw, circle, fill=yellow!50, minimum size=0.6cm] (circleB) at (8.75, 2.5) {};
    \draw[->] (rectB3.north) -- (circleB.west);
    \foreach \i in {4,5,6} {
        \draw[->] (circleB.east) -- (rectB\i.north);
    }
    \node at (8, 0) {\textbf{(b)} Sharing activation in a single large region.};
\end{tikzpicture}
}
\end{center} 
\caption{Schematic illustration of layer grouping: multiple small blocks versus a single large block. The colors indicate \textcolor{blue!80}{\textbf{compute-}} and \textcolor{orange!80}{\textbf{sharing-mode}}. The number of layers shown is for illustrative purposes; both approaches can accommodate a variable number of layers.}
    \label{fig:grouping_sharing_layers}
\end{figure*}
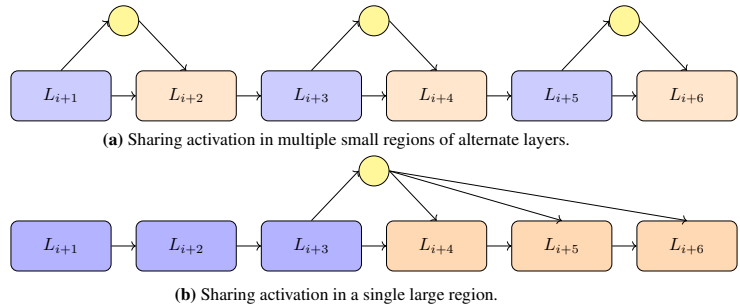

\textbf{Single vs. Multiple Sharing Block.}
Recent approaches of attention, $Q,~K$ and $K,~V$ sharing demonstrate two distinct methodologies for applying activation sharing across multiple layers. One line of research advocates for small activation-sharing blocks~\cite{ying2021lazyformer,rajabzadeh2024echoatt} while others argue that such fine-grained partitioning adds unnecessary complexity and computational overhead per block and favors instead a large activation-sharing block in the deeper end of the model~\cite{sun2024you}. To examine the trade-off between simplicity and computational efficiency, we conduct a comparative experiment with an equal number of layers in activation reuse mode for TinyLLaMA. One setup employs three groups of two activation-reusing layers, while the computationally equivalent model with a single block arranges three activation-reusing layers sequentially, as schematically depicted in Fig.~\ref{fig:grouping_sharing_layers}. Results in Table~\ref{tab:grouping_sharing_layers} consistently show superior performance with a single large activation-sharing block, attributed to deeper-layer sharing. This trend holds across training settings with and without EPAS. Adhering to Occam’s razor, we favor the simplicity of a single large sharing block, as additional complexity yields no clear advantage.

\vspace{-4pt}

\begin{wraptable}{r}{0.55\textwidth}
\centering
\begin{adjustbox}{width=0.54\columnwidth}
\begin{tabular}{l|ccccc}
\hline
\textbf{Model} &\textbf{PIQA} & \textbf{WG} &\textbf{BoolQ} & \textbf{OBQA} & \textbf{HS}\\
\hline 
MB (w/o EPAS) & 72.25 & 58.01 & \textbf{49.02} & 36.20 & 60.15 \\    
\hline
SB (w/o EPAS) & \textbf{73.50} & \textbf{60.14} & 48.41 & \textbf{37.40} & \textbf{60.84}\\
\hline
\hline 
MB (w/ EPAS) & \textbf{73.67} & 57.93 & 60.06 & 36.40 & 58.28\\    
\hline
SB (w/ EPAS) & 73.45 & \textbf{58.41} & \textbf{60.31} & \textbf{37.00} & \textbf{58.55}\\
\hline
\end{tabular}
\end{adjustbox}
\caption{Results comparing using a single large sharing block (SB) vs. multiple small sharing blocks (MB)
}
\label{tab:grouping_sharing_layers}
\end{wraptable}

\section{Related Works}
\label{sec:related}
Accelerating deep neural networks has been a tremendous research attention resulting in various approaches, such as focused on training data~\cite{li2022automated,hajimolahoseini2023swiftlearn,ataiefard2024skipvit}, model parameters~\cite{hajimolahoseini2023training,ahmed2023speeding,hajimolahoseini2024accelerating}, computation graph~\cite{fanreducing,zhang2020accelerating,he2024matters} and activation sharing~\cite{javadi2023gqkva}. Notably, training efficiency has focused on progressively freezing some layers~\cite{brock2017freezeout}, progressive stacking~\cite{gong2019efficient}, progressive layer drop~\cite{zhang2020accelerating}, and progressively increasing training overload in parallel to model growth~\cite{li2022automated}. Conversely, efficient inference has focused on model pruning~\cite{cheng2024survey}, low-rank adaptation~\cite{hu2022lora,liu2024dora}, and distillation approaches~\cite{yang2024survey} where recent works claimed to find progressive low-rank decomposition~\cite{hajimolahoseini2024single} and step by step distillation~\cite{hsieh2023distilling} as superior than single step low-rank decomposition or distillation. 

Cross layer activation sharing to leverage the redundancy observed in representation of deeper layers of transformers has recently emerged as a promising direction for enhancing model efficiency. Prominent cross-layer activation sharing methods include sharing attention scores, queries and keys ($Q,~K$), or keys and values ($K,~V$) across layers. For example, LazyFormer~\cite{ying2021lazyformer} introduced "lazy blocks," where the first layer in a group computes the attention scores and shares them with subsequent layers. ShareAttn~\cite{liao2024beyond} later extended this idea by using a single large block during inference following pretrained model analysis. LISA~\cite{mu2024cross} further enhanced attention score sharing by adding transformations to better align shared representations. However, these methods are incompatible with efficient block-factorized attention mechanisms like Flash-Attention~\cite{dao2022flashattention,dao2023flashattention}. An alternative strategy focused on sharing $Q,~K$ across layers and employed distillation-based techniques to train the model~\cite{rajabzadeh2024echoatt}. Cross-layer $K,~V$ sharing, on the other hand, reduces inference memory requirements by eliminating the need for separate K/V projections in each layer \cite{brandon2024reducing,sun2024you,dialameh2025echo}. 

While prior activation sharing approaches improved inference speed, there remains challenge in training these models and closing the gap in accuracy. Prior works in progressive training methods have been largely restricted to layer stacking or dropping and have not explored activation sharing. In contrast, EPAS introduces progressive activation sharing by bridging the ideas of the two areas to address the training challenges and severe accuracy drop of state-of-the-art activation sharing models.



\section{Conclusion}
\label{sec:conclusion}
EPAS presented an efficient transformer training algorithm incorporating a switchable decoder layer. It integrates cross-layer activation sharing as a generic property of the model and training process rather than a drop in modification of trained model. This design showcased a comprehensive solution for improving efficiency in both training and inference, balancing optimization to maintain performance while reducing computation. By applying EPAS during continual pretraining, pretrained models can be efficiently converted into activation-sharing models with adaptable computational budgets. Empirical evaluations demonstrated that EPAS enhances training and inference throughput with accuracy similar to the baseline. These findings underscore the potential of redundancy-aware training and inference as a scalable approach to optimizing transformer models. Furthermore, EPAS holds promise for extending to post-training as well as vision and speech domains, which merit further exploration in future research.

\printbibliography[heading=subbibintoc]

\end{document}